\documentclass[arxiv,article,accept,pdftex,moreauthors]{mdpi} 

\firstpage{1} 
\usepackage{color}
\pubvolume{0}
\issuenum{0}
\articlenumber{0}
\pubyear{2023}
\copyrightyear{2023}
\datereceived{2023} 
\daterevised{2023 } 
\dateaccepted{2023} 
\datepublished{2023} 
\hreflink{https://arxiv.org} 

\Title{Photovoltaic Panel Defect Detection Based on Ghost Convolution with BottleneckCSP and Tiny Target Prediction Head Incorporating YOLOv5} 

\TitleCitation{Photovoltaic Panel Defect Detection Based on Ghost Convolution with BottleneckCSP and Tiny Target Prediction Head Incorporating YOLOv5}

\Author{Longlong Li $^{1,2}$, Zhifeng Wang $^{1,}$*\orcidA{} and Tingting Zhang $^{2}$}

\AuthorNames{Longlong Li, Zhifeng Wang, Tingting Zhang}

\AuthorCitation{Li, L.; Wang, Z.; Zhang, T.}

\address{%
$^{1}$ \quad Faculty of Artificial Intelligence in Education, Central China Normal University, Wuhan 430079, China \\ 
$^{2}$ \quad School of Computer Science and Engineering, Northwest Normal University, Lanzhou 730070, China }

\corres{Correspondence: zfwang@ccnu.edu.cn}

\abstract{{Photovoltaic (PV)} panel surface-defect detection technology is crucial for the PV industry to perform smart maintenance. Using computer vision technology to detect PV panel surface defects can ensure better accuracy while reducing the workload of traditional worker field inspections. However, multiple tiny defects on the PV panel surface and the high similarity between different defects make it challenging to {accurately identify and detect such defects}. This paper proposes an approach named Ghost convolution with BottleneckCSP {and a tiny} target prediction head incorporating YOLOv5 (GBH-YOLOv5) for PV panel defect detection. {To ensure better accuracy on multiscale targets, the BottleneckCSP module is introduced to add a prediction head for tiny target detection to alleviate tiny defect misses,  using Ghost convolution to improve the model inference speed and reduce the number of parameters.} First, the original image is compressed and cropped to enlarge the defect size physically. Then, the processed images are input into GBH-YOLOv5, and the depth features are extracted through network processing based on Ghost convolution, the application of the BottleneckCSP module, and the prediction head of tiny targets. Finally, the extracted features are classified by a Feature Pyramid Network (FPN) and a Path Aggregation Network (PAN) structure. Meanwhile, {we compare our method with state-of-the-art} methods to verify the effectiveness of the proposed method. The proposed PV panel surface-defect detection network improves the mAP performance by at least 27.8\%.}


\usepackage{xcolor}
\usepackage{xpatch}
 
\makeatletter
\def\changeBibColor#1{%
  \in@{#1}{Hesham,DesignOP,GhostNetMF,Jumaboev,Wei_2022,Mellit_2022,segovia_ramirez_fault_2022,kirubakaran_infrared_2022,azkona_detection_2022,lu_generative_2022,kurukuru_machine_2022,bu_electrical_2022}
  \ifin@\else\normalcolor\fi
}
 
\xpatchcmd\@bibitem
  {\item}
  {\changeBibColor{#1}\item}
  {}{\fail}
 
\xpatchcmd\@lbibitem
  {\item}
  {\changeBibColor{#2}\item}
  {}{\fail}
\makeatother
\begin{document}

\setcounter{section}{0} 



\section{Introduction}

The high public demand for attention to environmental issues has become an essential indicator in the energy sector;~to deal with the ecological problems that affect all human beings, reduce greenhouse gas emissions, and~avoid the catastrophic consequences of climate change, countries around the world and the World Health Organization have developed relevant policies~\cite{Ko_2019}. The~terms ``carbon peak'' and ``carbon neutral'' are the most critical energy and environmental policies to cope with the global warming problem. To~achieve the ``double carbon'' goal, renewable energy, represented by photovoltaic power generation, is undoubtedly the main force {\cite{Wei_2022}}. As~an essential part of the development of the PV industry, the~fault detection of PV panels is of great significance in promoting the development of PV energy~\cite{Almalki_2022}. With~the development of artificial intelligence, {the intelligent} detection of PV panel faults is {becoming a feasible and promising solution.} Using machine vision techniques to identify surface defects in PV panels has become an essential technical basis for building intelligent PV inspection systems~\cite{Waqar_Akram_2022,Zeng2022}. Inspired by the success of deep learning in data mining~\cite{Wang2023,Lyu2022,Li2023a}, computer vision ~\cite{Wang2021,Zeng2021c,Wang2022ac,Zeng2020,Li2023,Zeng2022,Wang2017,Zeng2022b,Wang2015a,Zeng2020a} and speech processing ~\cite{Wang2021m,Zeng2021a,Wang2022t,Zeng2022a,Wang2020h,Zeng2021b,Wang2018a,Zeng2018,Wang2015b}, deep learning techniques can significantly improve detection efficiency, provide solutions for the competent inspection of PV power plants, and~guide power plants' operation and maintenance procedures~\cite{Guerriero_2016,Wang2022ac}.

The current processing techniques for PV panel images are mainly divided into two categories~\cite{Rahman_2021}. The~first category is the traditional machine learning methods, {which mostly rely on} manually designed extractors and require the manual construction of complex recognition relationships {\cite{Mellit_2022}}, {and their generalization ability and robustness could be better} \cite{AbdulMawjood_2018, Jiang_2022, Lyu2022}. YOLO and {Region-CNN (R-CNN)} algorithms, represented by deep learning techniques, are another class of methods that rely mainly on learning a large number of samples to obtain a deep dataset feature representation with better generalization ability and robustness~\cite{Mantel_2019, Wang2021}. Inspired by the previous research, we use YOLOv5 as the primary network framework, which is fast while maintaining good accuracy~\cite{Di_Tommaso_2022, DBLP:journals/corr/abs-2108-11539}.

This paper also introduces the BottleneckCSP module and the Ghost convolution mechanism, which help the model to obtain more information about the characteristics while maintaining the detection speed. YOLOv5s are used to detect five types of defects on the surface of PV panels: broken, hot\_spot, black\_border, scratch, and~no\_electricity. {At the same time, this paper compares five detection frameworks within the same family as YOLOv3: the~bipartite target detection methods Faster-RCNN and Mask-RCNN, the~traditional machine learning method SVM, and~Single Shot MultiBox Detector.} The contributions of this paper can be summarized as~follows:
\begin{itemize}
\item	{To the best of our knowledge, we are the first to apply the YOLOv5 structure to tackle the task of detecting defects on PV panels. This study utilizes the fast inference speed and high detection accuracy of YOLOv5 to obtain a combination of detection speed and accuracy on the PV Multi-Defect dataset, which enables accurate and rapid detection of various types of defects in PV panels {and significantly reduces the missed detection of minor defects.}}
\item   {According to the PV panel defect detection task, the~structure of YOLOv5 is improved and innovated in this paper. Firstly, the~semantic depth information of PV panel images is obtained using the BottleneckCSP module, improving detection accuracy. Secondly, the~added detection head for tiny targets alleviates the negative impact of drastic scale changes and improves the small target misdetection phenomenon. On~this basis, Ghost convolution is introduced instead of conventional convolution, and~we call this structure GBH-YOLOv5, which can perform the PV panel defect detection task well. The~implementation codes of this research are released at 
 \url {https://github.com/CCNUZFW/GBH-YOLOv5} (accessed on 23 December 2022).}
\item	{In this paper, a~new database dedicated to PV defect detection is constructed, which includes 5 types of defect targets and 1108 images with an image size of \mbox{600 × 600 pixels.} There are 886 images in the training set and 222 in the validation set. Moreover, the~database is publicly released to promote the field at the following links: \url{https://github.com/CCNUZFW/PV-Multi-Defect} (accessed on 23 December 2022).}
\item   {By comparing this method with five state-of-the-art methods, the~proposed PV panel surface defect approach has improved the mAP by at least 27.8\%, and~the single image detection time consumed is in the same order of magnitude, balancing detection accuracy and detection speed. It provides significant advantages in identifying various types of defects on the surface of PV panels.}
\end{itemize}

The remainder of the paper is structured as follows: Section~\ref{sec2} describes PV panel defect detection and the related studies on YOLO. Section~\ref{sec3} describes the defect detection process and the network framework, and~in Section~\ref{sec4}, comparison and ablation experiments are performed. Finally, the~conclusions of the article are stated in Section~\ref{sec5}.

\section{Related~Work} \label{sec2}
This section presents two parts of the related work: (2.1) the current state of research on PV panel defect detection and (2.2) the development of target detection based on the YOLO~algorithm.
\subsection{PV Panel Defect~Detection}
With the progress in energy structures, photovoltaic power generation, considered the most promising approach, is developing rapidly and {playing a significant role in} energy security, national income, public health, and~environmental protection. As~an essential component of a PV power generation system, PV panels are subject to challenging working environments and prone to faults, which affect the operation and lifetime of the entire PV system. Therefore, the~fault detection of PV panels is the key to improving PV systems' efficiency, reliability, and~lifecycle. There are three mainstream detection methods: image processing-based methods, electrical detection-based methods, and~machine learning-based~methods.
 
  \textbf{(1). Image processing-based methods
}:  Among the image processing-based methods, various imaging solutions exist depending on the different characteristics of the panels. In~thermal imaging, an~infrared camera is used to scan the PV array, which is suitable for {inspecting large PV plants.} The ultrasonic imaging inspection method is {used primarily for detecting cracks before the production of PV modules;} {an electroluminescence imaging} (EL image) solution is a unique image presented by the panel at a specific voltage, which is more expensive to detect. In~conclusion, imaging solutions rely on the various types of image features produced by PV panels under different techniques to determine their fault type. In~\cite{Serfa_Juan_2020}, by~varying the modulation of the injected current, the~panel image was made to exhibit certain features that allowed the detection of different types of shunt faults. {In~\cite{segovia_ramirez_fault_2022}, the~authors have verified that high accuracy fault identification is possible by performing thermal imaging analysis of PV panels and using radiation sensors. V. Kirubakaran~et~al.~\cite{kirubakaran_infrared_2022} use a thermal imaging system combined with image processing to record PV panel failure points.}

  \textbf{(2). Electrical detection-based methods
  }: Electrical detection-based methods include basic current--voltage measurement techniques, advanced {Climate-Independent Data (CID)} plans, and~power loss detection methods that enable fault detection and classification. Electrical detection methods diagnose specific faults based on different electrical characteristics. In~\cite{Vergura_2015}, the~{Time Domain Reflector (TDR)} technique is used to locate PV module faults based on the delay between the injected and reflected signals. In~addition, the~output voltage and current of the PV panel string are measured to identify possible faults in advance. {A.L.~et~al.~\cite{azkona_detection_2022} constructed a model for local defect and thermal breakdown detection of PV panels based on thermal images and IV curves.}

  \textbf{(3). Machine learning-based methods
  }: Since the performance and efficiency of PV cells are subject to various conditions, many problems are difficult to define in specific projects. Machine learning techniques can overcome these difficulties very well due to their self-learning nature, making them widely used in this type of detection~\cite{Zeng2022b}. In~\cite{Aouat_2021}, the~authors used grayscale cogeneration matrices to extract image features generated by infrared imaging techniques to monitor defects in panel modules. In~\cite{Mantel_2019}, the~authors used support vector machines (RBF kernel) and random forest algorithms to construct detection models to obtain the desired detection accuracy in the electroluminescence dataset. {Jumaboev~et~al. verified the feasibility of deep learning techniques in photovoltaic inspection by using several deep learning models~{\cite{Jumaboev}}}. {F.L.~et~al.~\cite{lu_generative_2022} proposed a semi-supervised anomaly detection model based on adversarial generative networks for PV panel defect detection. In~\cite{kurukuru_machine_2022},  an automatic detection method for optoelectronic components was proposed based on texture analysis and supervised learning for the processing of infrared images. Chiwu Bu~et~al.~\cite{bu_electrical_2022} used LDA and QDA supervised learning algorithms for the processing and defect identification of photovoltaic panel thermographic sequences.}

\subsection{Target Detection Based on~YOLO}
The YOLO algorithm is a one-stage target detection method proposed by Joseph Redmon. He converted the object detection problem into a regression problem by discarding the branching phase of candidate box extraction in the two-stage target detection algorithm and completing the determination of the entire category and the regression of the position in a single network~\cite{Redmon_2016}. 

In YOLOv2, the~authors introduced anchor boxes and batch normalization to improve the problem of the low detection accuracy of the v1 model~\cite{Redmon_2017}. YOLOv3 built a new Darknet with 53 residual networks based on YOLOv2 and passed feature pyramid networks for multiscale fusion prediction, which improved the detection accuracy of small and heavy targets~\cite{DBLP:journals/corr/abs-1804-02767}. YOLOv4 constructed a simple target detection model, which reduced the training threshold of this algorithm~\cite{DBLP:journals/corr/abs-2004-10934}. YOLOV5 constructed five models of YOLOv5, N/S/M/L/X, based on the scaling of different channels and the model size~\cite{DBLP:journals/csse/SuYWD23}. 

In the second half of 2022, YOLOv6 and YOLOv7 were released almost simultaneously. The~Meituan technical team introduced the RepVGG structure in YOLOv6, which was more adaptable to GPU devices and  simplified the adaptation work during engineering deployments~\cite{DBLP:journals/corr/abs-2209-02976}. YOLOv7 used module re-referencing and a dynamic label assignment strategy, which made it faster and more accurate in the range of 5 FPS to 160 FPS and exceeded known target detectors~\cite{DBLP:journals/corr/abs-2207-02696}.

In the PV panel surface-defect detection system used in this paper, the~C3 module in the network structure of YOLOv5s is replaced with the BottleneckCSP module, and~the number of detection heads is increased so that the grid acquires more feature information and improves the detection capability for small targets. In~addition, the~conventional convolution is replaced with Ghost convolution to reduce the model's parameters and the computational effort in the inference~process.

\section{Methods} \label{sec3}
{The task of finding defects in PV panels has two characteristics: first, it must precisely pinpoint the fault; second, it must accurately define the defect attributes. The~models in the YOLOv5 series are quite good at localization, and~YOLOv5s is also lighter. Compared to other deep learning techniques, YOLOv5s is able to strike a balance between detection accuracy and speed, making it a good choice for this purpose.}

\begin{figure}[H]
\begin{adjustwidth}{-\extralength}{0cm}
\centering
\includegraphics[width=18.25cm]{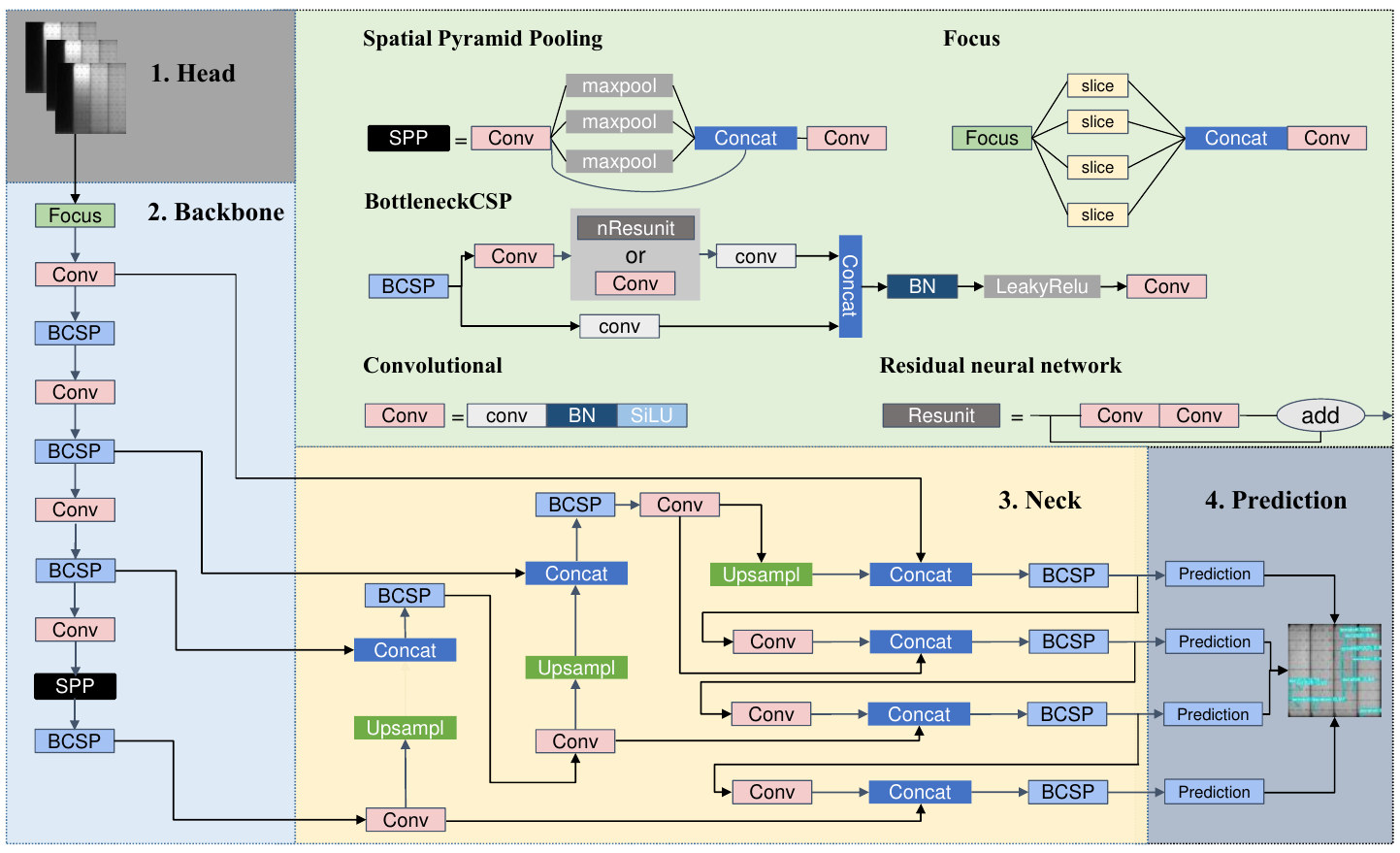}
\end{adjustwidth}
\caption{{The framework of PV panel surface-defect detection network. Using the BottleneckCSP module in the backbone network and neck network ensures that deeper semantic information of PV panels can be extracted, and~ a tiny target detection head is added at the prediction end so that the PV panel missed detection phenomenon is improved. Replacing the traditional convolution with Ghost convolution can guarantee the accuracy of PV panel detection without losing~speed.}\label{fig1}}
\end{figure}

The network structure of this {research is implemented} by improving the network structure of YOLOv5s, {named GBH-YOLOv5}, as~shown in Figure~\ref{fig1}.

{The network consisted of four parts: the head, backbone, neck, and~prediction.} The head was the input, and~the Mosaic data enhancement and adaptive image scaling were used to expand the data samples, enrich the test data set, and~increase the robustness of the network on the one hand; on the other hand, the~problem of the uneven distribution of the small target dataset was solved. By~setting different anchor frames and constantly updating the difference between the prediction frame and the labeled frame, {the adaptive anchor frame updated the network parameters and independently calculated the optimal anchor frame value to learn the feature information about the target better.} The backbone network was implemented by Focus, SPP, and~BottleneckCSP, based on Ghost convolution, and~the input 960 × 960 pixel image was sliced to obtain a 480 × 480 × 12 feature map; then, a \mbox{480 × 480 × 32} feature map was obtained after convolution. The~PV panel surface-defect detection network used a Focus structure to reduce the number of network layers and parameters, improve the forward and reverse computation speed, and~ensure no information loss in downsampling. In~order to extract more feature information, this network used a BottleneckCSP module with two deep convolutions. {To reduce the redundant feature} information brought by the BottleneckCSP module, Ghost convolution was used instead of the traditional convolution to reduce the parameters and the computation of the model inference process. {The prediction} was the output side, where CloU Loss was used as the loss function of the bounding box to solve the problem of the non-overlap between the labeled and predicted boxes; the Non-Maximum Suppression (NMS) mechanism was used to enhance the recognition ability of multiple targets and fuzzy targets. {In order to ensure that the scratch} and defective break were accurately recognized, a~detection head was added to the output of the network to realize the detection of tiny~targets.

\begin{figure}[H]
\includegraphics[width= 8 cm]{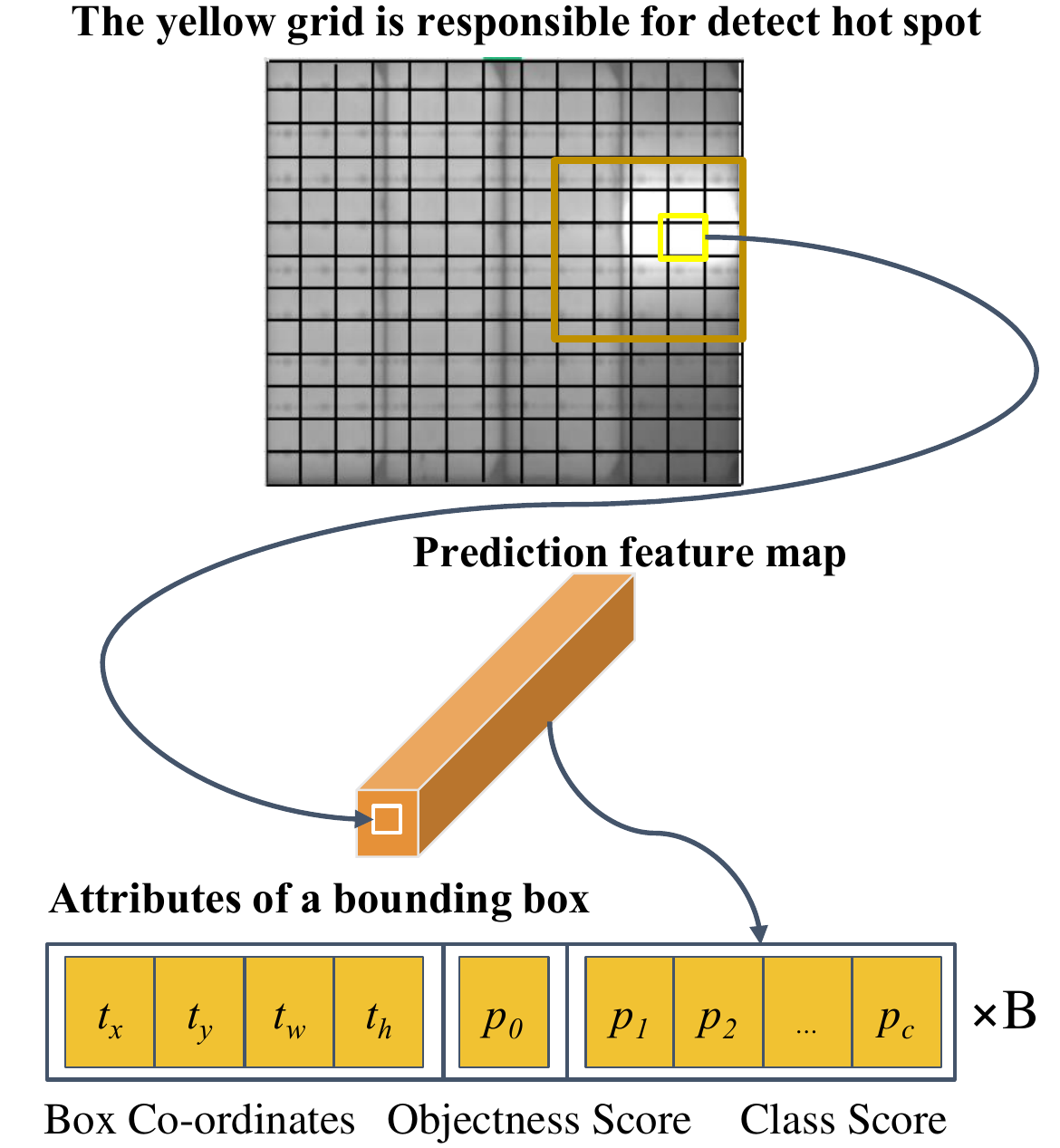}
\caption{Basic detection~idea.\label{fig2}}
\end{figure}

When performing detection, the~network divides the input image into $S\times S$ grids (grid detection method)~{{\cite{Redmon_2016}}}. If~the target's center point is inside a grid, {that grid is responsible} for the object detection grid. The~detection idea is shown in Figure~\ref{fig2}. When the grid detects the object, it outputs a bounding box, and~each bounding box consists of five parameters, i.e.,~four coordinate parameters and one confidence parameter. $t_x$ and $t_y$ denote the coordinates of the center point of the bounding box, and~$t_w$ and $t_h$ {represent the width and height of the bounding box,} respectively. The~confidence level indicates whether the current bounding box includes the object to be detected and its accuracy~\cite{Wang2022t,Zeng2021a}.

\subsection{PV Panel Defect Detection~Process}
\begin{figure}[H]
\begin{adjustwidth}{-\extralength}{0cm}
\centering
\includegraphics[width=18.1cm]{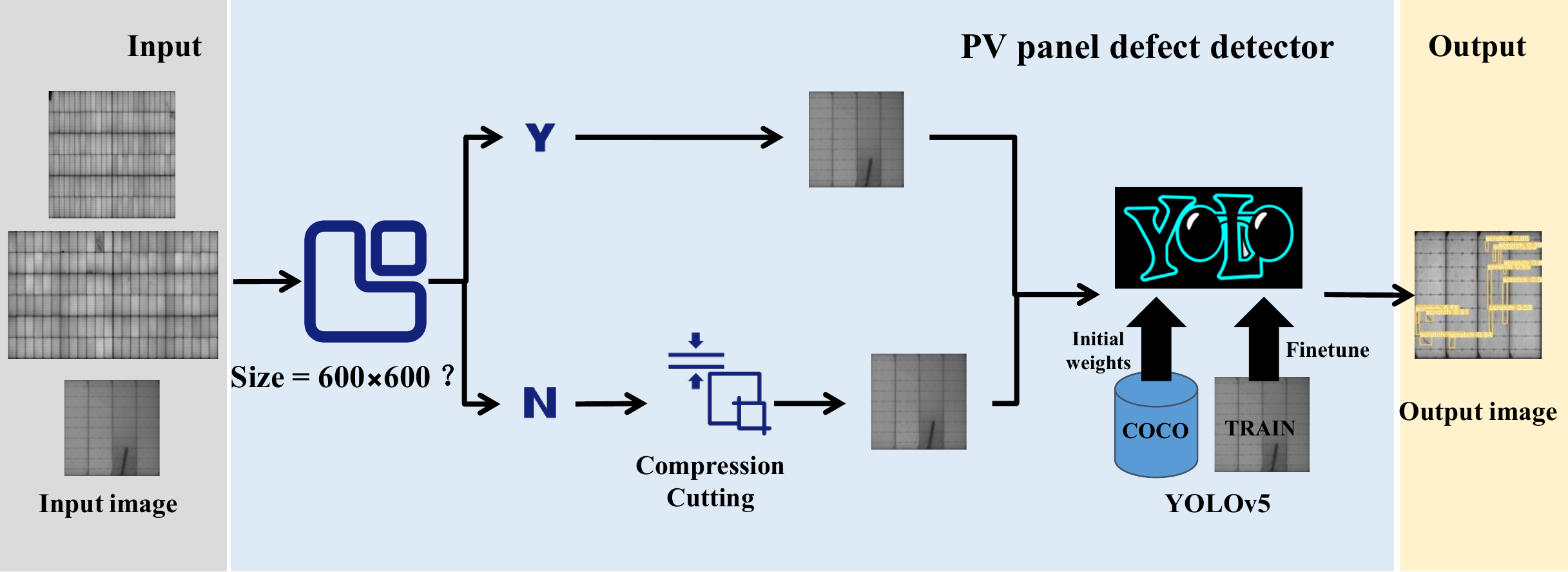}
\end{adjustwidth}
\caption{PV panel defect detection~process.\label{fig3}}
\end{figure}

The detection method proposed in this paper was composed of three processing modules, {mainly used for surface-defect} detection on the PV panels, as~shown in Figure~\ref{fig3}.

\textbf{(1) Input module
}: This module input the captured images into the PV panel defect detector, which has no requirement for the input size of the~images.

\textbf{(2) PV panel defect detector module
}: First, the~size of the input image was checked, and~for images whose  size was not 600 × 600 pixels, a~cropping and compression process was performed, which physically enlarged the panel defect and reduced the negative sample information. Then, pretrained weights based on the COCO dataset were used to train in the modified YOLO~network.

\textbf{(3) Output module
}: Defect detection was performed on the resulting images using the YOLOv5~network.

\unskip

\unskip

\subsection{BottleneckCSP~Module}
The BottleneckCSP was mainly used to extract deep semantic information from images and fuse the feature maps of different scales to enrich the semantic information. The~role of its primary structure, bottleneck, combined shallow-level feature maps with deep-level feature maps downward by summation to ensure that the detector maintained good accuracy on multiscale targets. {After mixing the CSP}, the~feature maps were integrated at the network's beginning and end to reflect the gradients' variability. It can be expressed as Equation~(1).
\begin{linenomath}
\begin{equation}
\left. \begin{array} { l } { y = F ( x _ { 0 } ) = x _ { k } } \\ { = H _ { k } ( x _ { k - 1 } , H _ { k - 1 } ( x _ { k - 2 } ) , H _ { k - 2 } ( x _ { k - 3 } ) , \ldots , H _ { 1 } ( x _ { 0 } ) , x _ { 0 } ) }. \end{array} \right. 
\end{equation}
\end{linenomath}

$H_k$ is the operator function of the $k^{th}$ layer, which usually consists of a convolution layer and an activation function. A~$y$ function was introduced to optimize each $H$ function.
\begin{linenomath}
\begin{equation}
 y = M ( x _ { 0 ^ { \prime } } , T ( F ( x _ { 0 ^ { \prime \prime } } ) ,
\end{equation}
\end{linenomath}
 where $x_0$ can be divided into two parts along the channel, the $T$ function truncates the gradient flow, and~the M function is used to mix the two parts. We obtained an information-rich feature map to retain and accumulate more features from different sensory \mbox{fields \cite{DBLP:conf/cvpr/WangLWCHY20}}.

\subsection{Prediction Head for Tiny~Targets}
We analyzed the PV Multi-Defect dataset and found a considerable proportion of tiny targets (scratches), so we added another prediction head to detect tiny targets. Combined with the other three prediction heads, the~four-head structure sufficiently alleviated the negative impact of the drastic scale transformation and thus mitigated the missed detection phenomenon. We added an anchor frame for small targets, enhanced the features from the second layer of the backbone network, and~finally added a prediction head for the second~layer.

\subsection{GhostConv~Module}
\begin{figure}[H]
\includegraphics[width= 8 cm]{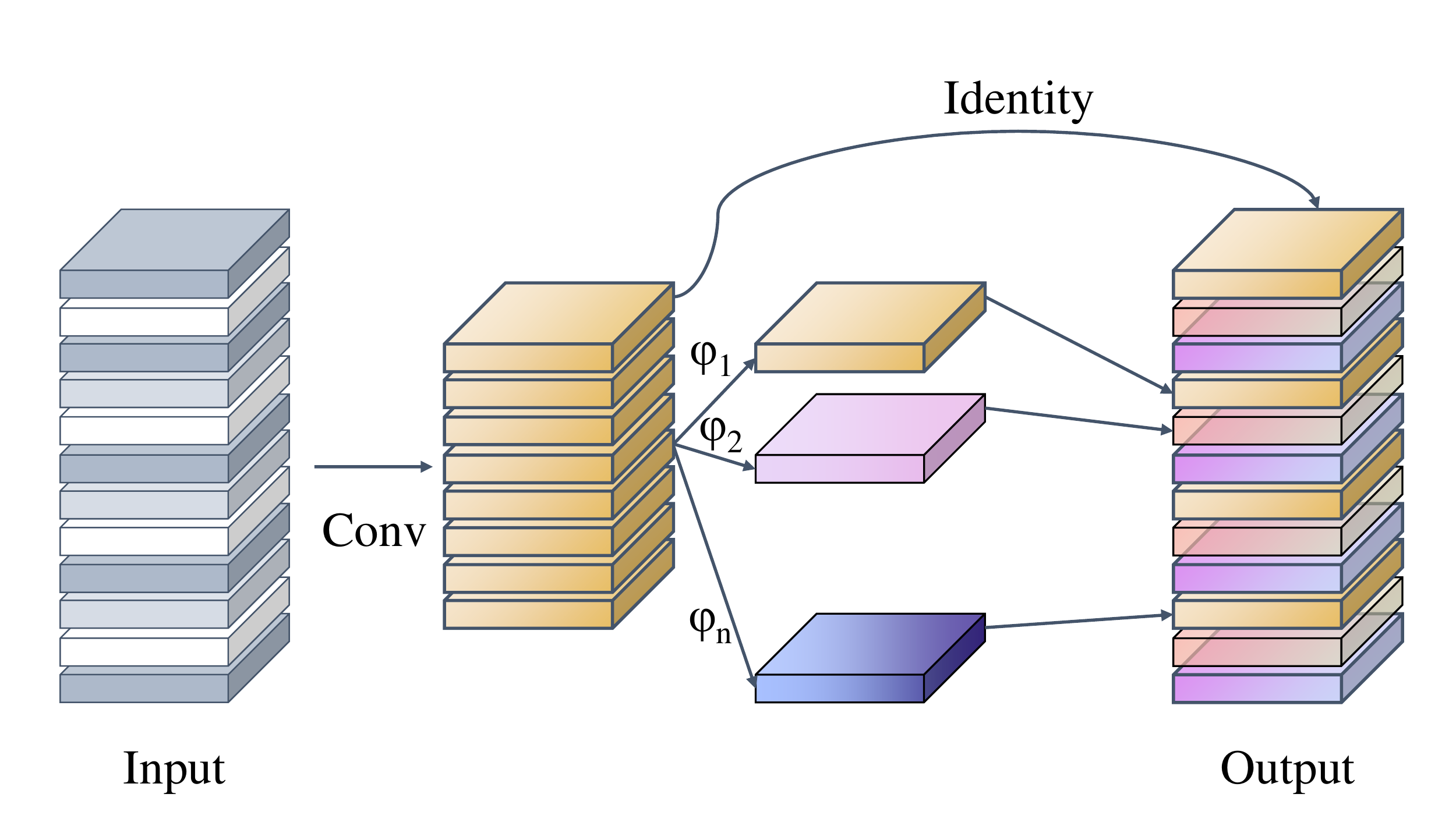}
\caption{ The Ghost convolution~process.\label{fig4}}
\end{figure}

{In Ghost convolution, only part of the feature map generated by conventional convolution was used to avoid the redundancy of the feature map.} Then, a~simple linear transformation was performed on this part of the feature map to achieve the effect of simulating conventional convolution~{{\cite{GhostNetMF}}}; the convolution process is shown in Figure~\ref{fig4}.

In this work, we used $X\in R^{c\times h\times w}$ to represent the input feature map, $c$ to represent the number of channels of the input feature map, and~$h$ and $w$ to express the height and width of the feature map, {respectively. The~conventional convolution is defined as Equation~(3).}
\begin{linenomath}
\begin{equation}
 Y=X*f+b .
\end{equation}
\end{linenomath}

In the above equation, $X\in R^{c\times h^{ \prime }\times w^{ \prime }}$ denotes a feature map with n channels output, and~$h^{ \prime}$ and $w^{ \prime}$ denote the height and width of the output feature map, respectively. $*$ denotes the convolution operation, the~convolution kernel size is $k*k$, b denotes the bias term, and~the regular convolution computation after ignoring the bias term is approximately equal to $h\times w\times c\times n\times w^{ \prime}\times h^{ \prime}$. In~the shallow layer of the network, $h^{\prime}$ and $w^{ \prime}$ {are more extensive,} and in the deeper layer, $n$ and $c$ are larger. Based on this feature, Ghost convolution was proposed, which consisted of two parts: the~regular convolution kernel {that outputs a small number of feature maps} and the generation of redundant feature maps in a lightweight linear transform layer, which can be expressed as
\begin{linenomath}
\begin{equation}
Y ^ { \prime } = X * f ^ { \prime } + b.
\end{equation}
\end{linenomath}

The above equation represents a conventional convolutional layer that outputs a small number of feature maps, where $Y^{ \prime}\in R^{h^{ \prime}\times w^{ \prime}\times m}$ represents the output feature and $f^{ \prime \in R^{c\times k\times k\times m}}$ represents the size of this convolutional kernel. The~number of channels of the output feature map is smaller than that of the conventional convolutional layer, i.e.,~m < n.
\begin{linenomath}
\begin{equation}
y _ { i j } = \phi _ { i , j } ( y _ { i } ^ { \prime } ).
\end{equation}
\end{linenomath}

Equation~(5) denotes the linear transformation layer that represents the generation of redundant feature maps, where $y_i$ denotes the m feature maps of $Y^{ \prime}${. Each feature map} in $Y^{ \prime}$ is subjected to a lightweight linear transformation operation $\phi_{i,j}(j=1,2,\dots,s)$ to obtain s feature maps. The~last linear transform is forcibly specified as a constant transform if the $d\times d$ convolution is used as the linear transform; so, $m$ feature maps are obtained after linear transformations of $m\times (s-1)$ feature maps. The~total computation using Ghost convolution is $(s-1)\times m\times h^{ \prime}\times w^{ \prime}\times k\times k$.
\section{Experimental Results and~Discussion} \label{sec4}
This section presents the dataset and describes the analysis and preprocessing; then, the~target detection field's baseline is discussed, and~comparison and ablation experiments are~described.
\subsection{{Dataset for~Experiments}}
\begin{figure}[H]
\includegraphics[width= 10 cm]{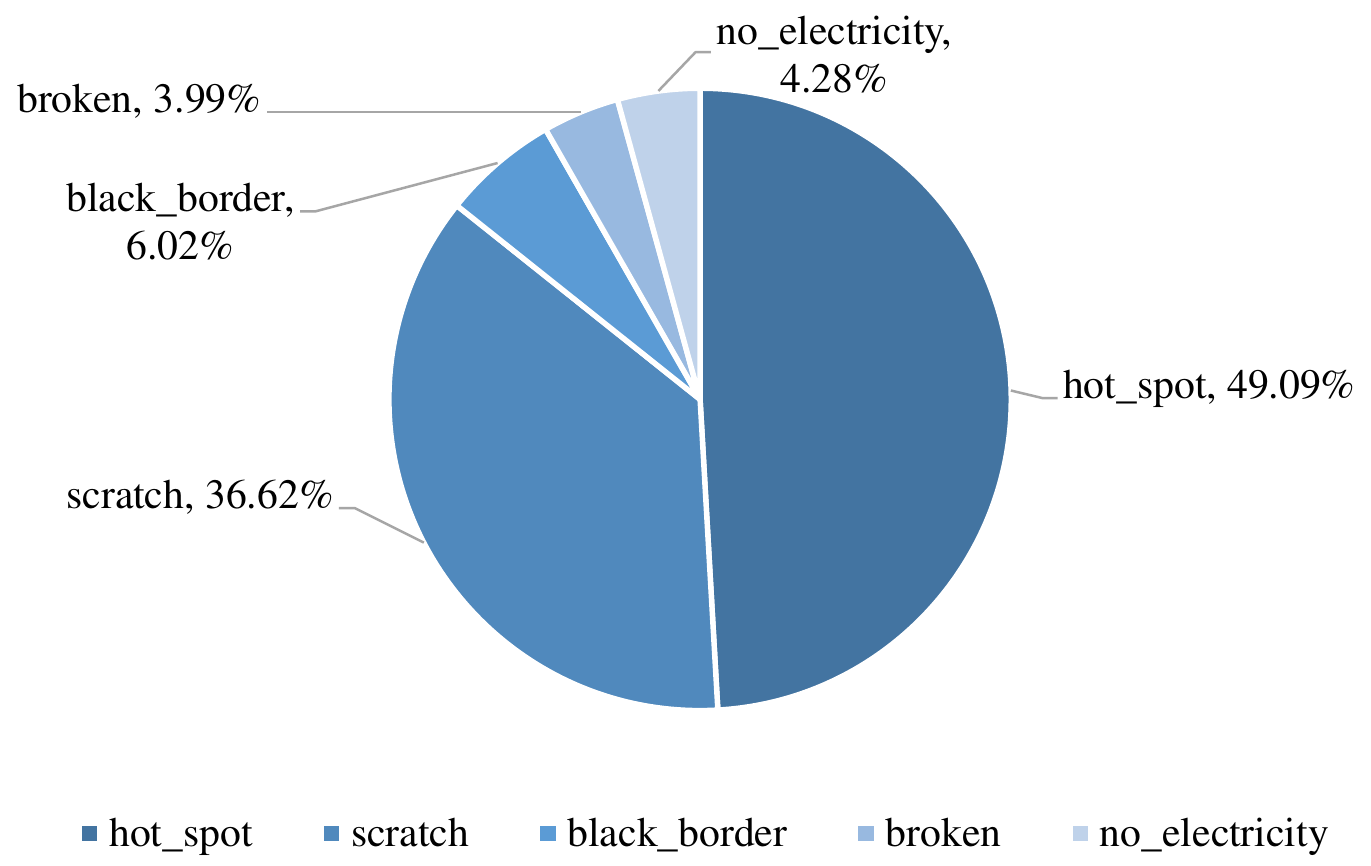}
\caption{ PV Multi-Defect dataset annotation~distribution.\label{fig5}}
\end{figure}

We constructed a publicly available dataset to verify the model's validity and named it the PV Multi-Defect dataset 
 (\url{https://github.com/CCNUZFW/PV-Multi-Defect} (accessed on 23 December 2022)). {The original images for this dataset were taken by the camera from photovoltaic modules with a physical size of 1.65 m × 0.991 m and a specification of 60~pieces. After~grayscale processing, the~images were uniformly cropped to 0.491~m~×~0.297~m in accordance with the distribution of defects, and~the images that did not demonstrate defects were manually removed. In total, 307 images, each measuring 5800 × 3504 pixels, were collected, as shown in Figure \ref{fig5}. In~this dataset, there were five common defect types, including broken cells, cells with prominent bright spots, cells with regularly shaped black or gray edges, cells with scratches, and~cells that were not charged and appeared black.} Table~\ref{tab1} shows examples of each defect~type. Figure \ref{fig6} shows the training and validation~losses.

The raw image data pixel size was 5800 × 3504 pixels, while the average size of the scratches was about 4 × 32 pixels. The~average length of the black edges was about \mbox{4 × 37 pixels}, the~average length of the broken edges was about 104 × 210 pixels, the~average size of the hot spots was about 152 × 210 pixels, and~the average size of the defective unpowered cell was 356 × 478 pixels. The~size of each type of defect was not uniform, and~the size of the defect was less than 0.08\% of the whole image. {If the original size image were used as the data for training, smaller targets would be detected with lower accuracy or even difficult to detect. Therefore, the~original image is preprocessed in this paper, and~the image size is changed to 600 $\times$ 600 pixels by compression and cropping operations. Figure
~\ref{fig7} compares the performance of each defect before and after data preprocessing ($mAP$). It can be found that preprocessing improves the training efficiency on the one hand and~makes the influence of some tiny target (scratch) noise reduced on the other hand, which effectively improves the accuracy of tiny target detection and increases the generalization ability of the training network.}

After preprocessing, we finally obtained 1108 defect images of the PV panel surface. We increased the size of the labeled boxes for each defect type, effectively improving the detection of small and fuzzy targets. The~final dataset was sequentially labeled using the LabelImg labeling software concerning the VOC2007 dataset format and converted to the XML format required for training. LabelImg is a labeling tool written in Python for deep learning image dataset production, which was used to label the information of the category name and the location of targets in the images. There were 886 images in the training set and 222 in the validation~set.

There were 4235 defective targets in 1108 images of the PV panel surface. Figure~\ref{fig5} shows that hot spots accounted for the highest percentage among the five types of defects, at~49.09\%. The~tiny target scratches accounted for 36.62\%, and~the blurred targets black border and broken cells accounted for 6.02\% and 3.99\%, respectively. 

\begin{table}[H]
\caption{Defect sample~diagram.\label{tab1}}
\footnotesize
	\begin{adjustwidth}{-\extralength}{0cm}
        \begin{tabular}{m{2cm}<{\centering}m{4.5cm}<{\centering}m{10.7cm}<{\centering} }
			\toprule
			\textbf{Name of Defect}	& \textbf{Description}	& \textbf{Image Style} \\
            \midrule
			broken & {Photovoltaic panels with \mbox{broken areas}} & \raisebox{0\height}{\includegraphics[width=0.9\linewidth]{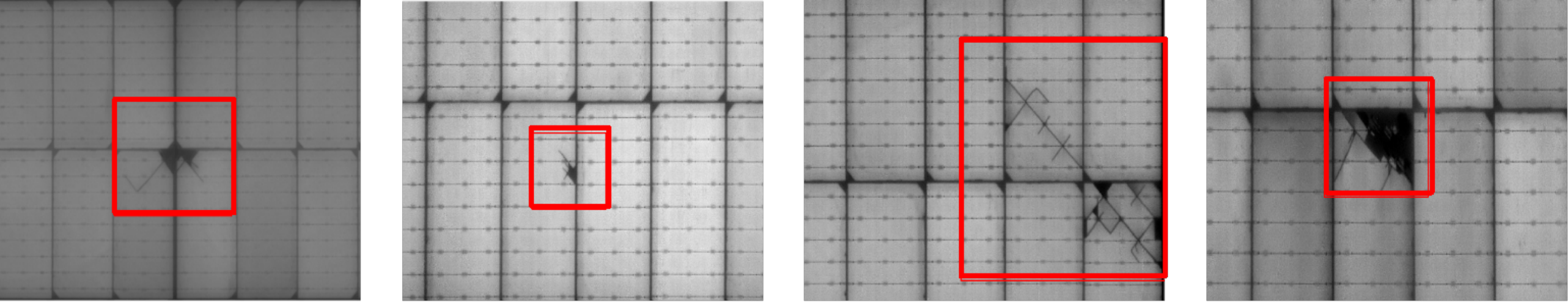}}	\\
			hot\_spot & {Photovoltaic panels have obvious bright spot areas}	& \raisebox{0\height}{\includegraphics[width=0.9\linewidth]{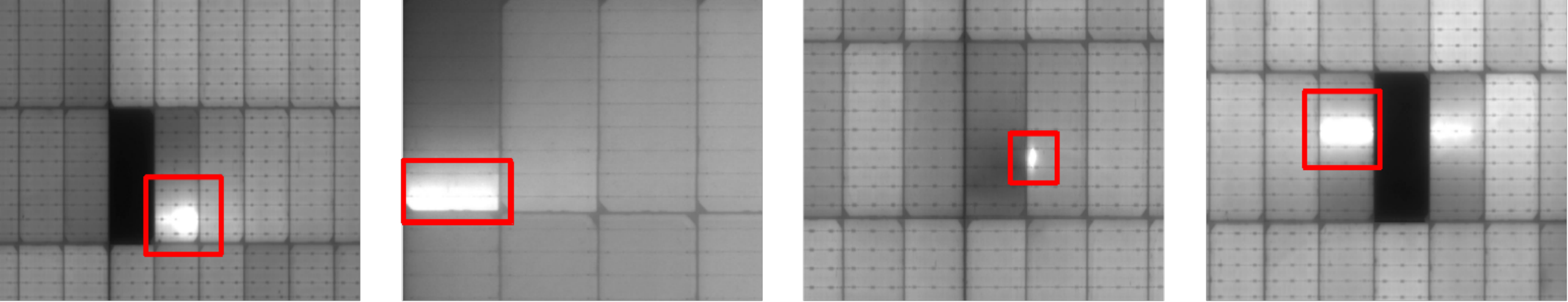}}
            \\
			black\_border & {Photovoltaic panels with black or gray border areas} & \raisebox{0\height}{\includegraphics[width=0.9\linewidth]{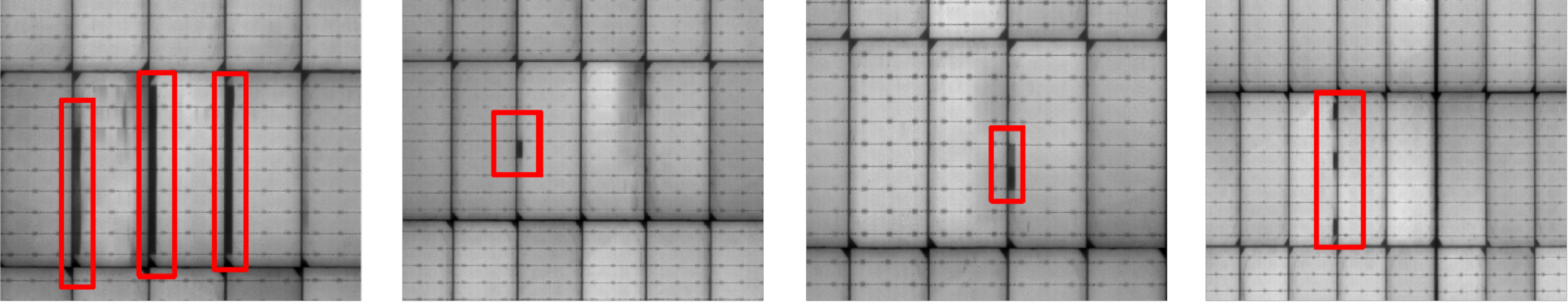}}	\\
            scratch & {Photovoltaic panels with \mbox{scratched areas}} & \raisebox{0\height}{\includegraphics[width=0.9\linewidth]{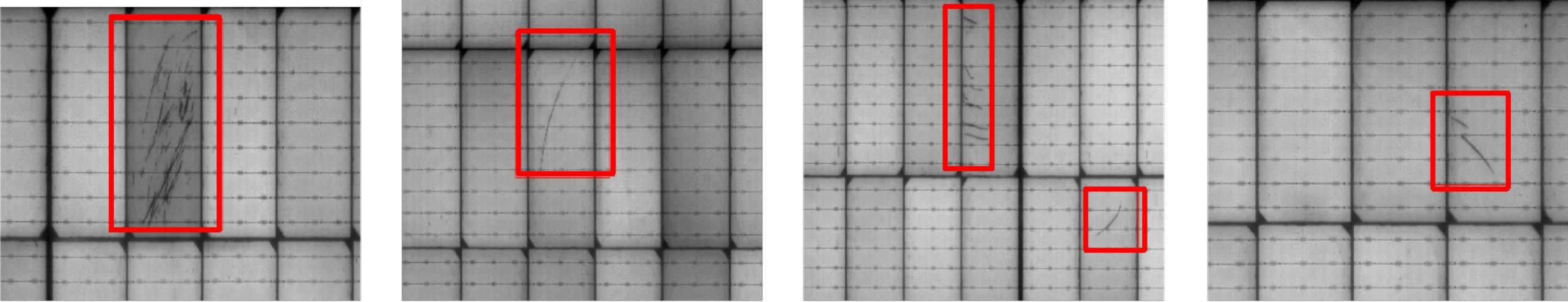}}	\\
            no\_electricity & {Photovoltaic panels have non-electricity and show black areas} & \raisebox{0\height}{\includegraphics[width=0.9\linewidth]{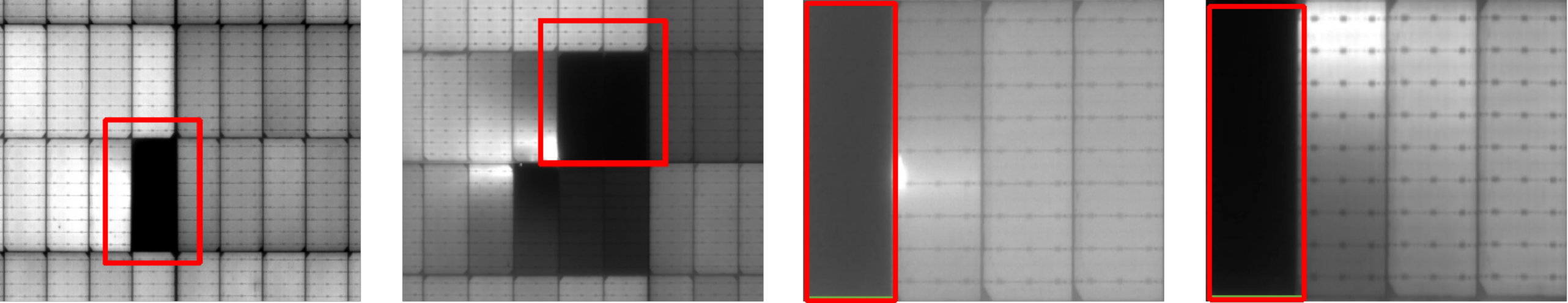}}	
            \\
			\bottomrule
		\end{tabular}
	\end{adjustwidth}
\end{table}
\unskip

\subsection{Baseline~Introduction}
{This work used the confusion matrix for supervised learning} as an evaluation metric. The~resulting evaluation criteria were the $Precision$, $Recall$, and~$mAP$ values~\cite{DBLP:journals/corr/abs-2004-10934}.

The $Recall$ is for the original sample and indicates how many positive sample cases were correctly predicted in the model, which is calculated in Equation~(6):
\begin{linenomath}
\begin{equation}
R = \frac { T P } { T P + F N } \times 100 \%.
\end{equation}
\end{linenomath}
where $R$ denotes the $Recall$ rate. $Precision$ is for the final prediction result, which indicates how many of the samples with positive predictions were really positive samples, calculated as Equation~(7).
\begin{linenomath}
\begin{equation}
P = \frac { T P } { T P + F P } \times 100 \%.
\end{equation}
\end{linenomath}
where $P$ denotes the $Precision$. In~the actual experiments, since both the $Recall$ and $Precision$ were maintained at a high level, a~parameter was needed to combine $Recall$ and $Precision$, i.e.,~{the performance of the algorithmic network is measured in terms of Mean Average Precision (mAP),} which applies to multitarget detection and is denoted as Equation~(8).
\begin{linenomath}
\begin{equation}
m A P = \frac { \sum _ { k = 1 } ^ { N } P ( k ) \Delta R ( k ) } { C }.
\end{equation}
\end{linenomath}

In the above equation, $N$ denotes the number of samples in the validation set, $P(k)$ denotes the magnitude of precision $P$ when $k$ targets are detected simultaneously, and~$\triangle R(k)$ denotes the change in $recall$ when the number of detected samples changes from $k - 1$ to $k$. $C$ denotes the number of classes of the~model.
\subsection{Experiment~Settings}
\begin{figure}[H]
\includegraphics[width= 13.5 cm]{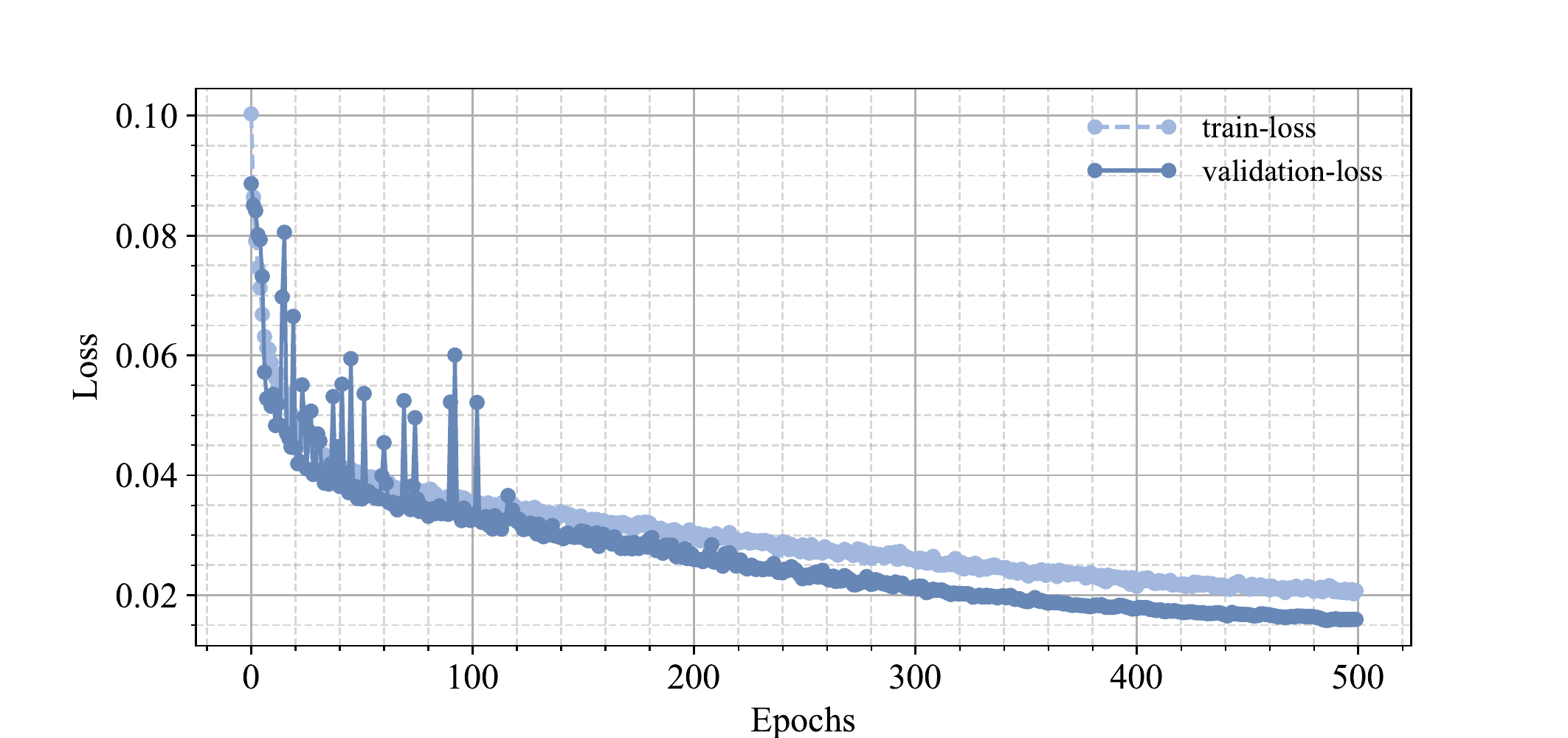}
\caption{{Training and validation losses for PV panel defect~detector.} \label{fig6}}
\end{figure}

\begin{figure}[H]
\includegraphics[width= 13.8 cm]{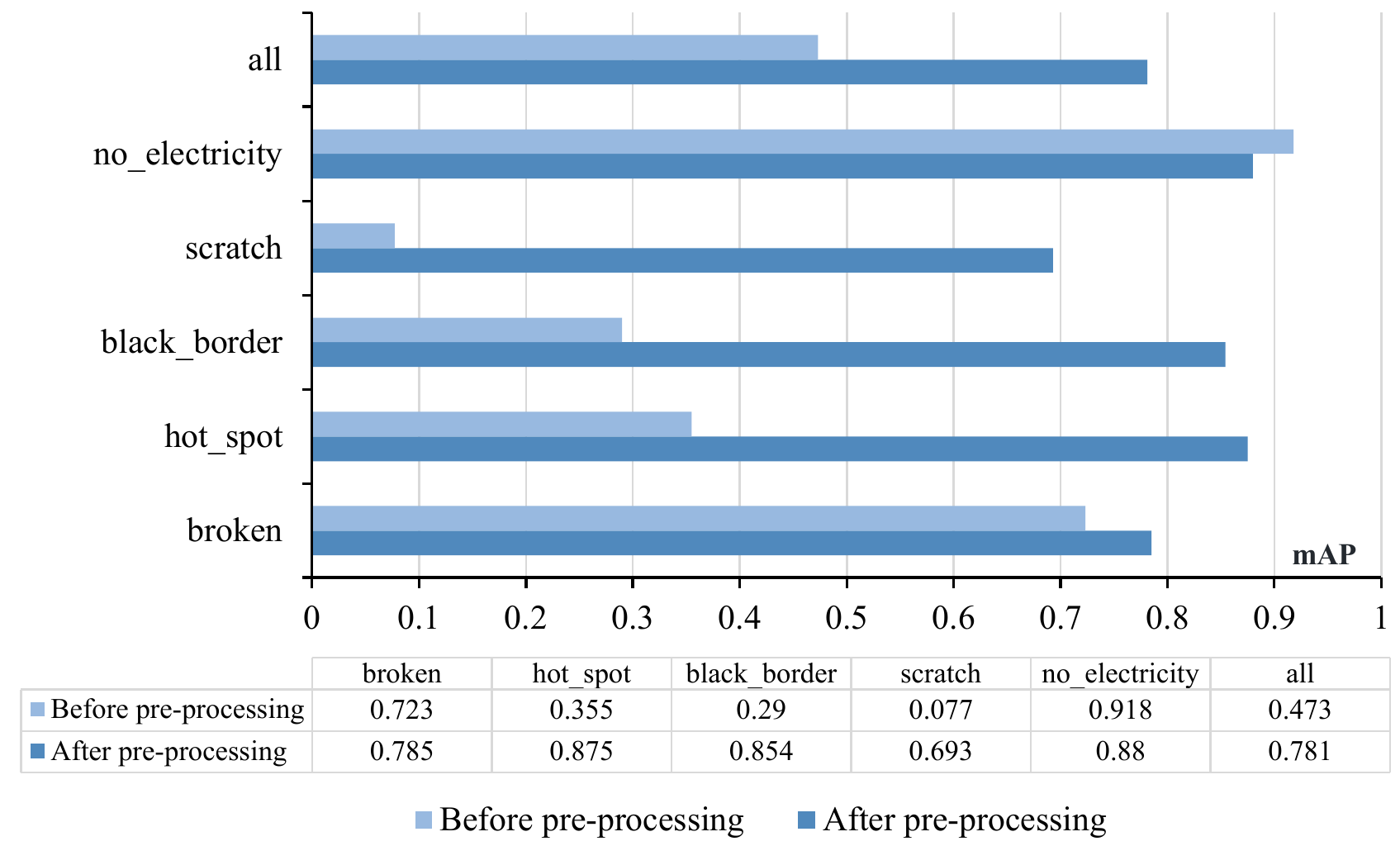}
\caption{{The 
 comparison of each defect's performance in the PV panel before and after data~pre-processing.} \label{fig7}}
\end{figure}

The experiments were conducted with 48 GB RAM and an RTX3090 graphics card with Pytorch and CUDA versions 1.8 and 11.1, respectively. {This study employs Adam as the optimizer according to the pre-training weight of the COOC data set to address the issue of inadequate data and increase learning speed and accuracy. The~batch size is 16, and~the learning rate is set at 0.001. As~demonstrated in Figure~\ref{fig6}, where the training and verification losses of the photovoltaic panel defect detector converge after 500 epochs, we monitor the loss of the bounding box by monitoring the loss of the photovoltaic panel defect detector in order to avoid overfitting. The~monitoring line demonstrates unequivocally that neither overfitting nor underfitting are present in the learned model. In~addition, the~validation set was held-out until the last test.}

{To make the model's performance optimal, the~$Recall$ and $Precision$ values during the training process are monitored in this paper. Table~\ref{tab2} shows their changes during the training of the model, and~according to the results, the~result of 500 training rounds is chosen as the model for subsequent test experiments in this paper.}
\subsection{Comparison with Other Methods on the Multi-Defect~Dataset}
{ We conducted a set of comparative experiments with the same database and experimental setup, and~all methods were retrained. In~addition, the~validation set was held-out until the last test. Comparing the methods proposed in this paper with the five techniques discussed in the literature, }{which are the YOLOv3-based method from Tommaso~et~al.~\cite{Di_Tommaso_2022}, the~Faster-RCNN-based method from Girshick~et~al.~\cite{girshick2015fast}, the~SVM-based method from Mantel~et~al.~\cite{Mantel_2019}, the~Mask-RCNN-based method from Almazroue~et~al.~\cite{Hesham}, and~the SSD-based method from Ren~et~al.~\cite{DesignOP},} the results showed that our performance on the Multi-Defect dataset was much better than the other models. The~specific mAP performance is listed in Table~\ref{tab3}.

\begin{table}[H] 
\caption{The 
 Performance of GBH-YOLOv5, {based on 95\% confidence~interval.}\label{tab2}}
\newcolumntype{C}{>{\centering\arraybackslash}X}
\begin{tabularx}{\textwidth}{CCC}
\toprule
\textbf{Epoch}	& \textbf{Recall (\%)}	& \textbf{Precision (\%)}\\
\midrule
0		& {0}			& {0}\\
50		& {80.8} $\pm$ {0.05}	& {62.9} $\pm$ {0.06}\\
100		& {79.6} $\pm$ {0.05}	& {73.1} $\pm$ {0.06}\\
150		& {87.0} $\pm$ {0.04}	& {77.9} $\pm$ {0.06}\\
200		& {84.6} $\pm$ {0.05}	& {88.1} $\pm$ {0.04}\\
250		& {86.3} $\pm$ {0.05}	& {93.0} $\pm$ {0.03}\\
300		& {91.7} $\pm$ {0.04}	& {90.1} $\pm$ {0.04}\\
350		& {91.9} $\pm$ {0.04}	& {90.7} $\pm$ {0.04}\\
400		& {93.3} $\pm$ {0.03}	& {93.5} $\pm$ {0.03} \\
450		& {93.8} $\pm$ {0.03}	& \textbf{94.5} $\pm$ \textbf{0.03}\\
500		&\textbf{96.4} $\pm$ \textbf{0.02}		& {93.3} $\pm${0.03}\\
\bottomrule
\end{tabularx}
\end{table}
\unskip

\begin{table}[H] 
\caption{The 
 mAP performance of different methods on the Multi-Defect dataset, {based on 95\% confidence~interval.}\label{tab3}}
\newcolumntype{C}{>{\centering\arraybackslash}X}

\begin{tabular}{m{6.5cm}<{\centering}m{6.53cm}<{\centering}}
\toprule
\textbf{Methods} & \textbf{mAP (\%)}\\
\midrule
Tommaso~et~al.~\cite{Di_Tommaso_2022} & {57.9} $\pm$ {0.07}\\

Girshick~et~al.~\cite{girshick2015fast} & {69.3} $\pm$ {0.06}\\

Mantel~et~al.~\cite{Mantel_2019} & {45.3} $\pm$ {0.07}\\

{Almazroue~et~al.~\cite{Hesham}} & {51.2} $\pm$ {0.06}\\

{Ren~et~al.~\cite{DesignOP}} & {30.8} $\pm$ {0.06} \\

Proposed GBH-YOLOv5		& \textbf{97.8} $\pm$ \textbf{0.02} \\
\bottomrule
\end{tabular}
\end{table}
\unskip
\subsection{Ablation~Studies}
We analyzed the performance of each component on the PV Multi-Defect dataset, and~the impact of each component is presented in Table~\ref{tab4}.


\begin{table}[H]
\caption{Ablation 
 Study of PV Multi-Defect datasets, {based on 95\% confidence~interval.}\label{tab4}}
	\begin{adjustwidth}{-\extralength}{0cm}
		\newcolumntype{C}{>{\centering\arraybackslash}X}
        \begin{tabular}{m{2.5cm}<{\centering}m{7.38cm}<{\centering}m{2cm}<{\centering}m{2.5cm}<{\centering}m{2cm}<{\centering}}
			\toprule
			\textbf{Methods} & \textbf{Description} & \textbf{mAP (\%)} &\textbf{Precision (\%)} &\textbf{Recall (\%)}\\
			\midrule
			YOLOv5s		& YOLOv5s		& {78.1} $\pm$ {0.06}		& {83.2} $\pm$ {0.05} & {73.4} $\pm$ {0.06}\\
            \midrule
			YOLOv5-1	& YOLOv5s + BottleneckCSP		& {94.2} $\pm$ {0.03}		& {88.2} $\pm$ {0.04} & {90.5} $\pm$ {0.04} \\
            \midrule
            YOLOv5-2	& YOLOv5s + BottleneckCSP + extra prediction		& {97.1}$\pm$ {0.02} & {93.4} $\pm$ {0.03} & \textbf{94.6} $\pm$ \textbf{0.03} \\
            \midrule
            GBH-YOLOv5	& YOLOv5s + BottleneckCSP + extra \mbox{prediction + GhostConv}		& \textbf{97.8} $\pm$ \textbf{0.02} & \textbf{96.4} $\pm$ \textbf{0.02} & {93.3} $\pm$ {0.02} \\
			\bottomrule
		\end{tabular}
	\end{adjustwidth}	
\end{table}
\unskip

\textbf{(1). The~effect of BottleneckCSP.
} After replacing the C3 residual module with the BottleneckCSP module, the~mAP values of the model were significantly improved. As~shown in Table~\ref{tab5}, {the performance on all defects was considerably enhanced}. The~problems of {tiny target missed detection and detection accuracy were} solved to a large extent. Therefore, the~improvement of the BottleneckCSP module was~excellent.

\begin{table}[H]
\caption{Performance 
 comparison of the YOLOv5 model for each category on the test~set.\label{tab5}}
	\begin{adjustwidth}{-\extralength}{0cm}
		\newcolumntype{C}{>{\centering\arraybackslash}X}
		\begin{tabularx}{\fulllength}{CCCCCC}
			\toprule
			\textbf{Methods} & \textbf{Broken (\%)} & \textbf{Hot\_Spot (\%)} &\textbf{Black\_Border (\%)} &\textbf{Scratch (\%)} &\textbf{No\_Electricity (\%)}\\
			\midrule
			YOLOV5s  & {78.5} $\pm$ {0.05}	& {87.8}  $\pm$ {0.04} & {85.4}  $\pm$ {0.02} & {69.3}  $\pm$ {0.06} & {88.0}  $\pm$ {0.04}\\
   
			YOLOv5-1 & {99.5} $\pm$ {0.01}	& {97.2}  $\pm$ {0.02} & {96.4} $\pm$ {0.02} & {95.6}  $\pm$ {0.02} & {97.7}  $\pm$ {0.02}\\ 
  
            YOLOv5-2 & {99.5}  $\pm$ {0.01}	& \textbf{98.4}  $\pm$  \textbf{0.02} & {96.7}  $\pm$ {0.02} & {96.4}  $\pm$ {0.02} & \textbf{98.9}  $\pm$  \textbf{0.01}\\
       
            GBH-YOLOv5 & \textbf{99.5}  $\pm$  \textbf{0.01} & {97.5}  $\pm$ {0.02} & \textbf{97.2}  $\pm$  \textbf{0.02} & \textbf{97.4}  $\pm$  \textbf{0.02} & {98.0}  $\pm$ {0.02}\\
		\bottomrule
		\end{tabularx}
	\end{adjustwidth}
\end{table}

\begin{figure}[H]
\begin{adjustwidth}{-\extralength}{0cm}
\centering
\includegraphics[width=18.25cm]{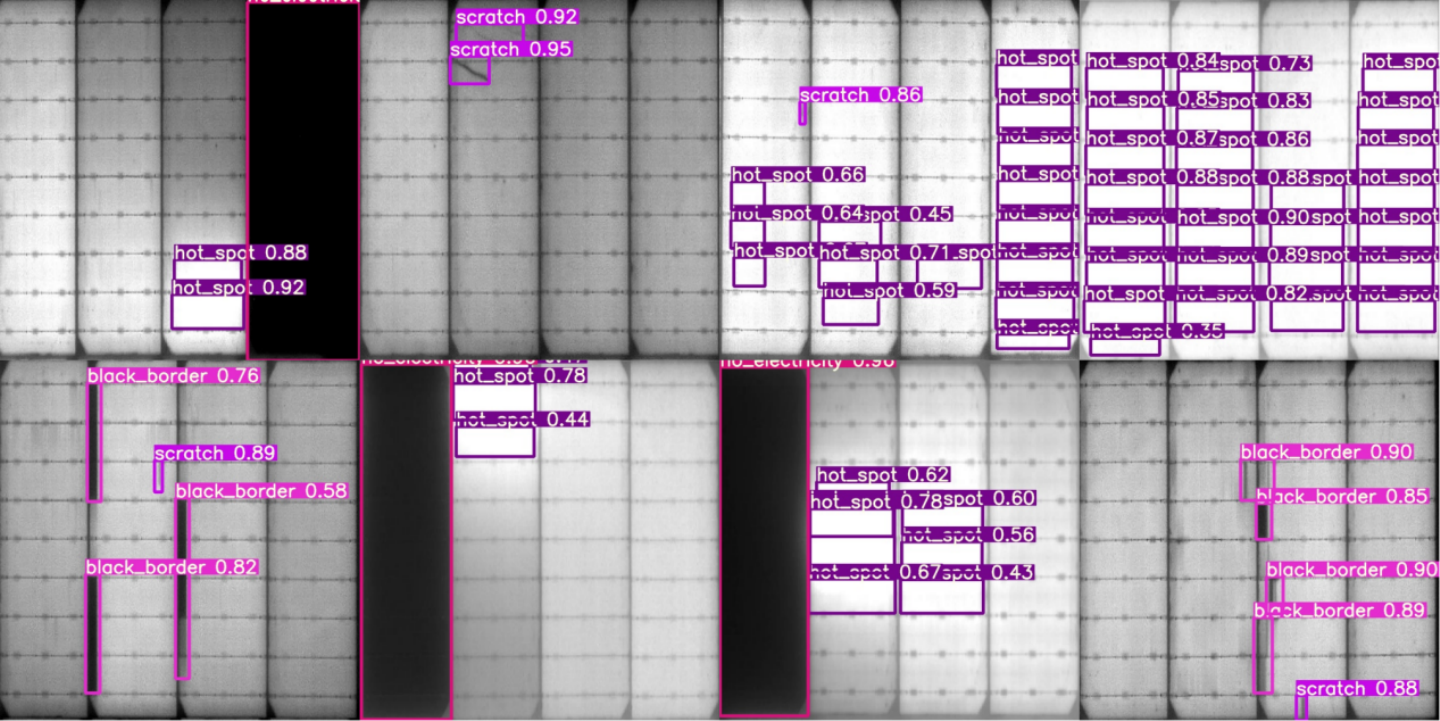}
\end{adjustwidth}
\caption{Surface defect identification results of PV~panel.\label{fig8}}
\end{figure}

\textbf{(2). The~effect of the extra prediction head.
} {As shown in Table~\ref{tab6}, the addition of the tiny target prediction head increased the number of network layers in YOLOv5-2 from 224 in YOLOv5s to 290. }Even though the computation and the number of parameters were increased, the~improvement in the mAP was significant. As~shown in Figure
~\ref{fig8}, GBH-YOLOv5 performed well in detecting tiny targets, so it was worth sacrificing some of the~computation.

\begin{table}[H] 
\caption{{The summary of the different models and the average elapsed time on the test~set.}\label{tab6}}
\newcolumntype{C}{>{\centering\arraybackslash}X}
\begin{tabular}{m{3.17cm}<{\centering}m{3cm}<{\centering}m{3cm}<{\centering}m{3cm}<{\centering}}
\toprule
{\textbf{Methods}} & {\textbf{Model Layers}} & {\textbf{Number of Parameters (\boldmath{$10^4$})}} &{\textbf{Average Time Consuming (s)}}\\
\midrule
{YOLOv5s}	 & {224} & {7.06} & {0.484}\\

{YOLOv5-1} & {228} & {7.15} & {0.658}\\

{YOLOv5-2} & {290} & {7.72} & {0.695}\\

{GBH-YOLOv5} & {270} & {7.24} & {0.587}\\
\bottomrule
\end{tabular}
\end{table}

 \textbf{(3). The~effect of GhostCov.
 } The use of Ghost convolution instead of regular convolution resulted in a reduction in the number of layers in the network of GBH-YOLOv5 from 290 to 270, and~the elapsed time on the test set was reduced by 0.108 s per image {on average while still} maintaining an excellent~mAP.

\textbf{(4). The~effect of the model ensemble.
} This paper lists the mAP values of {the four models} for different categories on the same test set. They are compared with the final integrated model (GBH-YOLOv5) in Table~\ref{tab5}, where the GBH-YOLOv5 achieved a relatively balanced result in maintaining accuracy and the time~duration.


\section{Conclusions} \label{sec5}
In this paper, we proposed an approach named Ghost convolution with BottleneckCSP and tiny target prediction head incorporating YOLOv5 (GBH-YOLOv5) for PV panel defect detection. To~ensure better accuracy on multiscale targets, {the BottleneckCSP module was introduced to add a prediction head for tiny target detection to improve the phenomenon of missed detection of tiny defects}, and~it used Ghost convolution to improve the model inference speed and reduce the number of parameters. First, the~original image was compressed and cropped to enlarge the defect size physically. Then, the~processed images were input into GBH-YOLOv5, and~the depth features were extracted through network processing based on Ghost convolution, the~application of the BottleneckCSP module, and~the prediction head of tiny targets. Finally, the~extracted features were classified by a Feature Pyramid Network (FPN) and a Path Aggregation Network (PAN) structure. Meanwhile, we compared our method {with state-of-the-art methods to verify its effectiveness.} The proposed PV panel surface-defect detection network improved the mAP performance by at least 27.8\%. {As the addition of modules makes the number of parameters of the model increase and the volume of the model become larger, the~selected dataset is grayscale processed and may generate some errors when detecting PV panel defects in a natural production environment. Possible future research directions include the use of lightweight networks with better real-time performance or the direct use of RGB images for PV panel defect detection.}



\newpage



\authorcontributions{Conceptualization, Z.W. and L.L.; methodology, Z.W. and L.L.; software, L.L. and T.Z.; validation, Z.W. and L.L.; formal analysis, Z.W. and L.L.; investigation, Z.W. and L.L.; writing---original draft preparation, L.L. and T.Z.; writing---review and editing, Z.W.; supervision, Z.W.; project administration, Z.W.; funding acquisition, Z.W. All authors have read and agreed to the published version of the~manuscript.}

\funding{The research work in this paper was supported by the National Natural Science Foundation of China (No. 62177022, 61901165, 61501199), the~Collaborative Innovation Center for Informatization and Balanced Development of K-12 Education by MOE and Hubei Province (No. xtzd2021-005), and~the Self-determined Research Funds of CCNU from the Colleges’ Basic Research and Operation of MOE (No. CCNU22QN013).}

\dataavailability{Data will be made available on reasonable request.} 

\acknowledgments{We would like to thank Chunyan Zeng 
 of Hubei University of Technology for her strong support of this~research.}

\conflictsofinterest{The authors declare no conflict of~interest.} 

\begin{adjustwidth}{-\extralength}{0cm}

\reftitle{References}
\bibliographystyle{mdpi} 
\bibliography{ref,Citations}

\end{adjustwidth}

\end{document}